\documentclass[letterpaper, 10 pt, conference, table]{IEEEtran}
\usepackage{etex}
\IEEEoverridecommandlockouts  

\usepackage[numbers,compress]{natbib}
\usepackage{multicol}
\usepackage[font={small}]{caption}

\usepackage{comment}
\usepackage{siunitx}
\usepackage{relsize}
\usepackage{ifthen}
\usepackage[colorinlistoftodos]{todonotes}

\usepackage[caption=false]{subfig}
\usepackage[]{caption}

\usepackage[vlined,ruled,linesnumbered]{algorithm2e}
\usepackage{graphics} %
\usepackage{rotating}
\usepackage{color}
\usepackage{enumerate}
\usepackage[T1]{fontenc}
\usepackage{psfrag}
\usepackage{epsfig} %
\usepackage{booktabs}
\usepackage{graphicx,url}
\usepackage{multirow}
\usepackage{array}
\usepackage{latexsym}
\usepackage{amsfonts}
\usepackage{amsmath}
\usepackage{amssymb}
\usepackage{xstring}
\usepackage[noend]{algorithmic}
\usepackage{multirow}
\usepackage{xcolor}
\usepackage{prettyref}
\usepackage{flexisym}
\usepackage{bigdelim}
\usepackage{breqn} %
\usepackage{listings}
\let\labelindent\relax
\usepackage{enumitem}
\usepackage{xspace}
\usepackage{bm}
\graphicspath{{./figures/}}
\usepackage{tikz}
\usetikzlibrary{matrix,calc}

\usepackage{mdwlist}

\let\itemize\undefined
\makecompactlist{itemize}{stditemize}

\usepackage{amsthm}
\newtheoremstyle{mystyle}%
  {}%
  {}%
  {\itshape}%
  {}%
  {\bfseries}%
  {.}%
  { }%
  {\thmname{#1}\thmnumber{ #2}\thmnote{ (#3)}}%
\theoremstyle{mystyle}
\newrefformat{prob}{Problem\,\ref{#1}}
\newrefformat{def}{Definition\,\ref{#1}}
\newrefformat{sec}{Section\,\ref{#1}}
\newrefformat{sub}{Section\,\ref{#1}}
\newrefformat{prop}{Proposition\,\ref{#1}}
\newrefformat{app}{Appendix\,\ref{#1}}
\newrefformat{alg}{Algorithm\,\ref{#1}}
\newrefformat{cor}{Corollary\,\ref{#1}}
\newrefformat{thm}{Theorem\,\ref{#1}}
\newrefformat{lem}{Lemma\,\ref{#1}}
\newrefformat{fig}{Fig.\,\ref{#1}}
\newrefformat{tab}{Table\,\ref{#1}}

\newcommand{\bdmath}{\begin{dmath}}
\newcommand{\edmath}{\end{dmath}}
\newcommand{\beq}{\begin{equation}}
\newcommand{\eeq}{\end{equation}}
\newcommand{\bdm}{\begin{displaymath}}
\newcommand{\edm}{\end{displaymath}}
\newcommand{\bea}{\begin{eqnarray}}
\newcommand{\eea}{\end{eqnarray}}
\newcommand{\beal}{\beq \begin{array}{ll}}
\newcommand{\eeal}{\end{array} \eeq}
\newcommand{\beas}{\begin{eqnarray*}}
\newcommand{\eeas}{\end{eqnarray*}}
\newcommand{\ba}{\begin{array}}
\newcommand{\ea}{\end{array}}
\newcommand{\bit}{\begin{itemize}}
\newcommand{\eit}{\end{itemize}}
\newcommand{\ben}{\begin{enumerate}}
\newcommand{\een}{\end{enumerate}}

\newcommand{\calR}{{\cal R}}

\newcommand{\setal}{~\emph{et~al.}\xspace}
\newcommand{\eg}{\emph{e.g.,}\xspace}
\newcommand{\ie}{\emph{i.e.,}\xspace}
\newcommand{\myParagraph}[1]{{\bf #1.}\xspace}
\newcommand{\M}[1]{{\bm #1}} %
\renewcommand{\boldsymbol}[1]{{\bm #1}}

\newcommand{\hide}[1]{}

\newcommand{\hiddenText}{{\color{gray} hidden text.}}
\newcommand{\hideWithText}[1]{\hiddenText}

\DeclareMathOperator*{\argmin}{arg\,min}

\newcommand{\Real}[1]{ { {\mathbb R}^{#1} } }

\newcommand{\SEthree}{\ensuremath{\mathrm{SE}(3)}\xspace}

\newcommand{\SOthree}{\ensuremath{\mathrm{SO}(3)}\xspace}

\newcommand{\MM}{\M{M}}

\newcommand{\MR}{\M{R}}

\newcommand{\MX}{\M{X}}

\newcommand{\vg}{\boldsymbol{g}}

\newcommand{\vv}{\boldsymbol{v}}
\newcommand{\vt}{\boldsymbol{t}}

\newcommand{\scenario}[1]{{\smaller \sf#1}\xspace}

\newcommand{\blue}[1]{{\color{blue}#1}}

\newcommand{\linkToPdf}[1]{\href{#1}{\blue{(pdf)}}}
\newcommand{\linkToPpt}[1]{\href{#1}{\blue{(ppt)}}}
\newcommand{\linkToCode}[1]{\href{#1}{\blue{(code)}}}
\newcommand{\linkToWeb}[1]{\href{#1}{\blue{(web)}}}
\newcommand{\linkToVideo}[1]{\href{#1}{\blue{(video)}}}
\newcommand{\linkToMedia}[1]{\href{#1}{\blue{(media)}}}
\newcommand{\award}[1]{\xspace} %

\newcommand{\kimera}{\scenario{Kimera}}
\newcommand{\kimeraMulti}{\scenario{Kimera-Multi}} %

\newcommand{\camp}{\scenario{Camp}}
\newcommand{\city}{\scenario{City}}
\newcommand{\euroc}{\scenario{EuRoc}}
\newcommand{\RBCD}{RBCD\xspace}

\newcommand{\kimeraVIO}{\scenario{Kimera-VIO}}
\newcommand{\kimeraSemantics}{\scenario{Kimera-Semantics}}
\newcommand{\LMO}{LMO\xspace}

\newcommand{\robot}{\calR} \graphicspath{{./figures/}}

\usepackage{float}
\usepackage{hyperref}
\usepackage{graphicx}
\usepackage{epstopdf}
\usepackage{epsfig}

\usepackage{xr}

\title{\huge{\kimeraMulti: a System for Distributed Multi-Robot  Metric-Semantic Simultaneous Localization and Mapping}}
\author{Yun Chang, Yulun Tian, Jonathan P. How, Luca Carlone
\thanks{Y. Chang, Y. Tian, J.P. How, and L. Carlone are with the Laboratory for 
Information \& Decision Systems, Massachusetts Institute of Technology, Cambridge, MA, USA, 
{\sf \{yunchang,yulun,jhow,lcarlone\}@mit.edu}
}
\thanks{This work was
partially funded by 
ARL Distributed and Collaborative Intelligent Systems and Technology Collaborative Research Alliance (DCIST CRA) under agreement W911NF-17-2-0181.}
}

\begin{document}

\maketitle

\begin{abstract}
We present the first fully distributed multi-robot  system for dense metric-semantic Simultaneous Localization and Mapping (SLAM).
Our system, dubbed \kimeraMulti, is implemented by a team of robots equipped with visual-inertial sensors, 
and builds a 3D mesh model of the environment in real-time, where each face of the mesh is annotated with a semantic label (\eg building,\,road,\,objects).
In \kimeraMulti,
each robot builds a local trajectory estimate and a local mesh using~\kimera~\cite{Rosinol20icra-Kimera}.
Then, when two robots are within communication range, they initiate a distributed place recognition and 
robust pose graph optimization protocol with a novel incremental maximum clique outlier rejection; 
the protocol allows the robots to improve their local trajectory estimates by leveraging inter-robot
loop closures. 
Finally, each robot uses its improved trajectory estimate to correct the local mesh using mesh deformation techniques. 
We demonstrate \kimeraMulti in photo-realistic simulations and real data.
\kimeraMulti 
(i)~is able to build accurate 3D metric-semantic meshes, 
(ii)~is {robust to incorrect loop closures} while requiring less computation than state-of-the-art distributed SLAM back-ends,
and (iii)~is efficient, both in terms of computation at each robot as well as communication bandwidth.
\end{abstract} %

\section{Introduction}
\label{sec:introduction}

Multi-robot systems have been the subject of continuous research by the robotics community due to 
their capability to sense and act over large-scale environments.  
This capability is key to improving efficiency and robustness in several applications, 
including factory automation, search \& rescue, intelligent transportation, planetary exploration, and
surveillance and monitoring in military and civilian endeavors. %
 
 In this paper, we are concerned with the problem of using a team of robots to gain situational awareness 
 over a large environment, under realistic constraints on communication bandwidth, local sensing, and 
 computation at each robot. In particular, our goal is to estimate a \emph{metric-semantic} 3D model of the environment that 
 describes the geometry of the scene the robots operate in (\eg presence and shape of obstacles), as well as its semantics, 
where the robots are tasked with annotating the obstacles with human-understandable labels  
in a given dictionary (\eg ``building'', ``road''). The last decade has seen a renaissance in single-robot metric-semantic SLAM, pioneered 
by works such as SLAM++~\cite{Salas-Moreno13cvpr} and SemanticFusion~\cite{McCormac17icra-semanticFusion}. 
Recent work includes systems that can build metric-semantic 3D models in real-time using a multi-core CPU, including  
Kimera~\cite{Rosinol20icra-Kimera} and Voxblox++~\cite{Grinvald19ral-voxbloxpp}. %
These research efforts are marking a steady transition from traditional geometric SLAM to 
\emph{spatial perception} (or \emph{Spatial AI}~\cite{Davison18arxiv-futureMapping}) approaches that aim at constructing high-level representations of the environment~\cite{Rosinol20rss-dynamicSceneGraphs}.
This ongoing success achieved by single-robot perception systems has not been fully harnessed in multi-robot systems.
This is partially due to the complexity of designing and deploying multi-robot systems, as well as the fact that building ``richer'' metric-semantic 
representations might negatively impact the amount of communication and computation needed 
for the robots to build such a model.
For these reasons, current multi-robot systems have mostly focused on geometric reasoning, with 
attention to communication aspects~\cite{Cieslewski18icra,Choudhary17ijrr-distributedPGO3D} 
or robustness~\cite{Mangelson18icra,Lajoie20ral-doorSLAM}, while disregarding dense semantics. 
This work advances the state of the art by developing the first system for distributed and dense metric-semantic SLAM.

\begin{figure}[t]
\centering
\includegraphics[width=0.99\columnwidth,trim=0mm 0mm 0mm 0mm,clip]{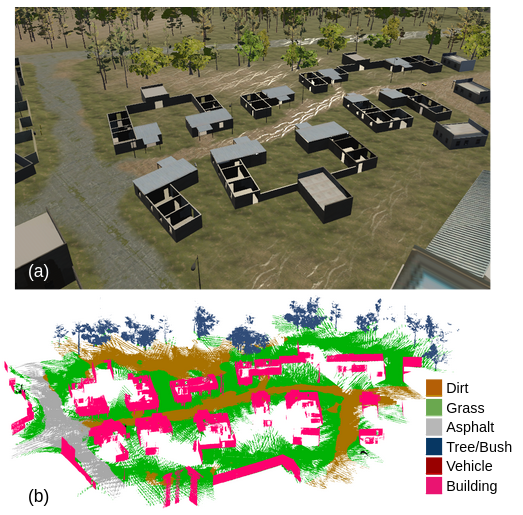}
\vspace{-3mm}
\caption{(a) \camp scene generated by the Unity-based DCIST multi-robot simulator~\cite{dcist}.
(b) Dense metric-semantic 3D mesh model generated by \kimeraMulti with three robots. \label{fig:cover} }
\vspace{-6mm}
\end{figure}

\myParagraph{Related Work}
Related work on Collaborative SLAM (CSLAM)
has focused on developing multi-robot SLAM front-ends (\eg to find inter-robot loop closures) 
and back-ends (\eg distributed pose graph optimization).
Inter-robot loop closures are critical to align the trajectories of the robots
in a common reference frame and to improve their trajectory estimates.
In a centralized setup, a common way to obtain loop closures is to use visual
place recognition methods, based on image or keypoint descriptors~\cite{Oliva01ijcv,Ulrich00icra,Oliva01ijcv,Ulrich00icra,Lowe99iccv,Bay06eccv,Sivic03iccv,Arandjelovic16cvpr-netvlad}. 
Recent works investigate \emph{distributed} inter-robot loop closure detection, where the
images are not collected at a single location and loop closures are found through local communication among the robots~\cite{tian2019resource,giamou2018talk,tian2018near,Cieslewski18icra,cieslewski2017efficient,van2018collaborative,Tardioli2015}
Centralized back-end approaches for multi-robot pose graph optimization (PGO) collect measurements
at a central station, which computes the trajectory estimates for all
robots~\cite{Andersson08icra,Kim10icra,Bailey11icra,Lazaro11icra,Dong15icra}. 
Since centralized approaches become impractical with large teams or in the presence of tight communication constraints, the research community has recently investigated \emph{distributed}  
PGO~\cite{Aragues11icra-distributedLocalization,tian2019distributed,tian2020asynchronous,fan2020majorization,cristofalo2020geod,Choudhary17ijrr-distributedPGO3D} and distributed factor graph solvers~\cite{Cunningham10iros,Cunningham13icra}.
Front-end and back-end algorithms have been also demonstrated in 
complete CSLAM systems such as \cite{Cieslewski18icra,Choudhary17ijrr-distributedPGO3D,Lajoie20ral-doorSLAM,Wang19arxiv}. 
While the multi-robot literature~\cite{SajadSaeedi2016MultipleRobotSL} has mostly focused on dense geometric representations 
(\eg occupancy maps~\cite{Choudhary17ijrr-distributedPGO3D}) or sparse landmark maps~\cite{Tchuiev20ral}, 
single-robot SLAM research is steadily moving towards systems that can build 
\emph{metric-semantic} maps~\cite{Tateno17cvpr-CNN-SLAM,Lianos18eccv-VSO,Dong17cvpr-XVIO,Behley19iccv-semanticKitti,McCormac17icra-semanticFusion,Zheng19arxiv-metricSemantic,Tateno15iros-metricSemantic,Li16iros-metricSemantic,McCormac183dv-fusion++,Runz18ismar-maskfusion,Runz17icra-cofusion,Xu19icra-midFusion,Rosinol20icra-Kimera,Grinvald19ral-voxbloxpp,Rosinol20rss-dynamicSceneGraphs}.
Related research efforts include systems building voxel-based models~\cite{McCormac17icra-semanticFusion,Zheng19arxiv-metricSemantic,Tateno15iros-metricSemantic,Li16iros-metricSemantic,McCormac183dv-fusion++,Runz18ismar-maskfusion,Runz17icra-cofusion,Xu19icra-midFusion}, 
ESDF and meshes~\cite{Rosinol20icra-Kimera,Rosinol19icra-incremental,Grinvald19ral-voxbloxpp},
 or 3D scene graphs~\cite{Rosinol20rss-dynamicSceneGraphs}.

\myParagraph{Contribution} 
We present \kimeraMulti, the first fully distributed system for multi-robot metric-semantic dense SLAM. 
The system enables a team of robots to build a semantically-annotated 3D mesh model of the environment in real-time 
by leveraging local sensing and computation, and intermittent communication
(Sec.~\ref{sec:overview}). 
In \kimeraMulti, each robot builds a local trajectory estimate and a local mesh by processing visual-inertial sensor data 
using~\kimera~\cite{Rosinol20icra-Kimera}. %
Then, when a pair of robots is within communication range, they initiate a distributed place recognition and 
robust PGO protocol with a maximum clique outlier rejection (Sec.~\ref{sec:trajEstimation}).
The distributed front-end and outlier rejection modules in \kimeraMulti are
similar to those of DOOR-SLAM~\cite{Lajoie20ral-doorSLAM}, except for our more efficient
\emph{incremental} maximum clique search heuristic
\cite{Pattabiraman15im-maxClique}. 
The distributed PGO back-end in \kimeraMulti is based on the Riemannian Block-Coordinate Descent (\RBCD) method of
Tian\setal\cite{tian2019distributed}, which has been shown to
outperform the 
Gauss-Seidel back-end in~\cite{Choudhary17ijrr-distributedPGO3D} (also used in~\cite{Lajoie20ral-doorSLAM} and~\cite{Cieslewski18icra}). 
As a result of \RBCD, the robots obtain an improved local trajectory estimate. 
After this distributed protocol is executed, each robot uses its 
improved trajectory estimate to correct the local mesh consistently with the 
loop closure (Sec.~\ref{sec:LMO}). We develop a real-time implementation of 
the mesh deformation approach of Sumner\setal~\cite{Summer07siggraph-embeddedDeformation} to correct the mesh.

We demonstrate \kimeraMulti in photo-realistic large-scale simulations and on
real data (Sec.~\ref{sec:experiments}).
The results show that \kimeraMulti (i) is able to build accurate 3D metric-semantic meshes, 
(ii) is robust to incorrect loop closures while requiring significantly less computation compared to~\cite{Lajoie20ral-doorSLAM,Mangelson18icra} when rejecting outliers,
and (iii) is efficient in terms of computation at each robot while achieving as much as 80\% reduction in communication compared to a typical centralized system. 

\section{{\large \kimeraMulti}: System Overview}
\label{sec:overview}

\begin{figure}[t]
	\centering
	\includegraphics[width=0.99\columnwidth,
	trim=10mm 10mm 35mm 25mm,clip]{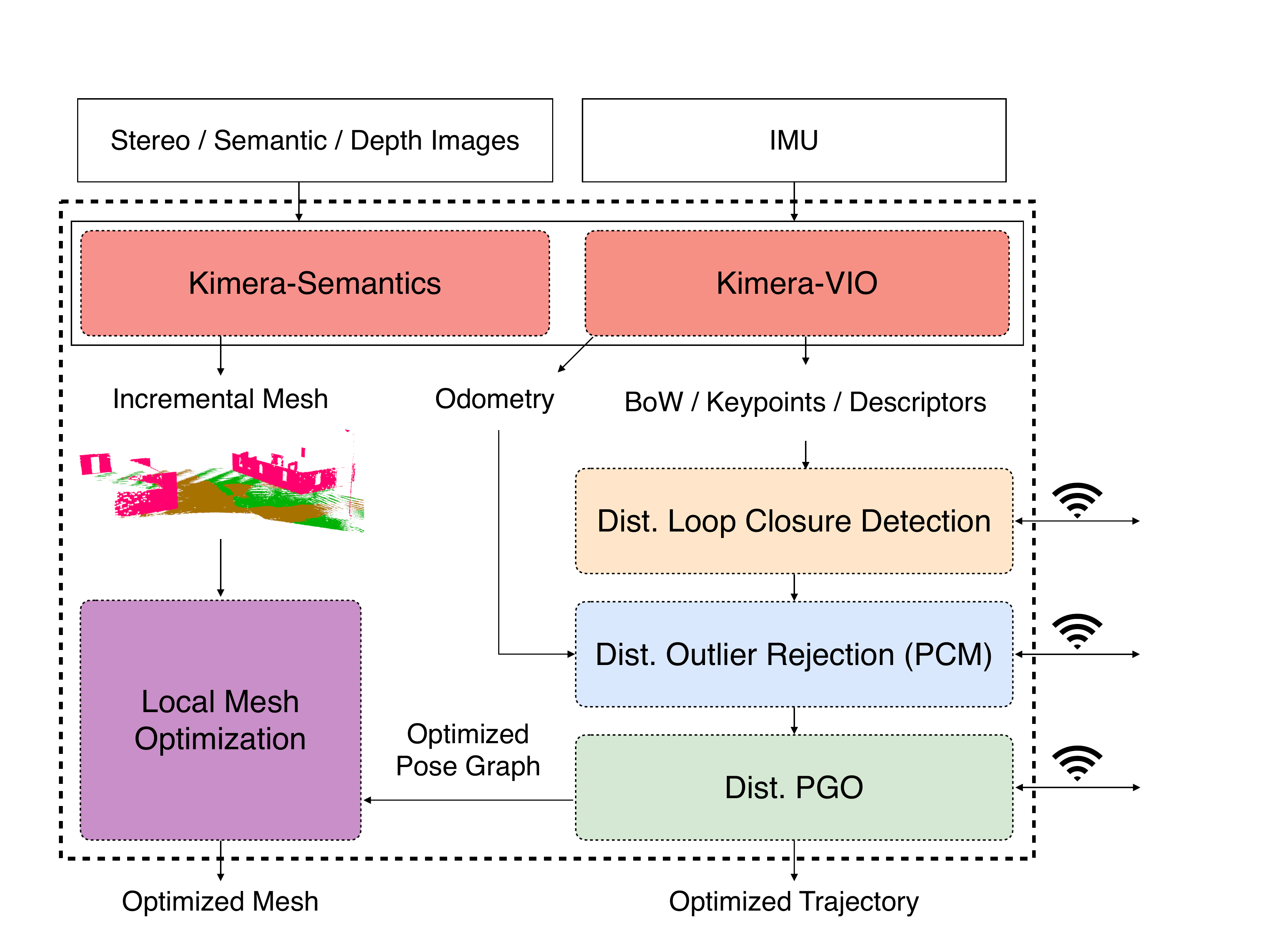}
	\vspace{-5mm}
	\caption{\kimeraMulti: system architecture. 
	Each robot runs \kimera (including \kimeraVIO and \kimeraSemantics)
	to estimate local trajectory and mesh. 
	Robots then communicate to perform distributed loop closure detection, outlier rejection, and PGO. 
	Given the optimized trajectory, each robot performs local mesh optimization.
	\label{fig:architecture} \vspace{-5mm}}
\end{figure}

The \kimeraMulti architecture --used by each robot in the team-- is displayed in Fig.~\ref{fig:architecture}.
\kimeraMulti includes five main modules: 
(i)~single-robot \kimera, 
(ii)~distributed loop closure detection, %
(iii)~distributed outlier rejection, %
(iv)~distributed PGO, 
and (v) local mesh optimization. %

{\bf Kimera}~\cite{Rosinol20icra-Kimera} is implemented by each robot $\robot_i$, and  estimates the local trajectory of the robot and a local mesh corresponding to the portion of the map 
seen by the robot. In particular, we use \kimeraVIO~\cite{Rosinol20icra-Kimera} as a visual-inertial odometry module, 
which processes raw stereo images and IMU data to obtain an estimate of the odometric trajectory of the robot. 
Moreover, we use \kimeraSemantics~\cite{Rosinol20icra-Kimera} to process depth images (possibly obtained by stereo matching) 
and a 2D semantic segmentation~\cite{GarciaGarcia17arxiv} and obtain a dense metric-semantic 3D mesh. 
\kimeraVIO also computes a Bag-of-Words (BoW) representation of each keyframe using ORB features and DBoW2~\cite{Galvez12tro-dbow}, 
which is used for distributed loop closure detection. 

{{\bf Distributed Loop Closure Detection (Sec.~\ref{sec:DLCD})}} is executed whenever two robots $\robot_i$ and $\robot_j$ are within communication range.
The robots exchange BoW descriptors of the keyframes they collected.
When the robots find a pair of matching descriptors 
--which typically correspond to 
poses observing the same place-- they estimate a relative pose 
using geometric verification as in~\cite{Lajoie20ral-doorSLAM}. 
The relative pose corresponds to a putative inter-robot loop closure between $\robot_i$ and $\robot_j$.

{\bf Distributed Outlier Rejection (Sec.~\ref{sec:DPCM})} is then used to vet the quality of the putative 
inter-robot loop closure using Pairwise Consistency Maximization (PCM)~\cite{Mangelson18icra}. We implement a fast incremental maximum clique
heuristic to find consistent inter-robot loop closures.

{\bf Distributed PGO (Sec.~\ref{sec:DPGO})} 
takes as input the inter-robot loop closures that pass the PCM check, along with the odometry measurements produced by \kimeraVIO, 
and finds the optimal trajectory given the measurements 
by both robots $\robot_i$ and $\robot_j$ using the \RBCD pose graph solver from~\cite{tian2019distributed}.

{\bf Local Mesh Optimization (Sec.~\ref{sec:LMO})} is executed after distributed PGO is complete, and 
is a local processing that deforms the mesh at each robot to enforce 
consistency with the trajectory estimate resulting from distributed PGO.

\kimeraMulti is 
implemented in C++ and uses the Robot Operating System~(ROS)~\cite{Quigley09icra-ros} 
as a communication layer between robots and between the modules executed on each robot.
The systems runs in real-time on a CPU and is modular, thus allowing modules to be replaced or removed.
For instance, the system can also produce a dense \emph{metric} mesh if semantic labels are not available, or only 
produce the optimized trajectory if the dense reconstruction is not required by the user.
\section{Distributed Trajectory Estimation}
\label{sec:trajEstimation}

This section describes the Distributed Loop Closure Detection, Distributed Outlier Rejection, and Distributed PGO 
modules in Fig.~\ref{fig:architecture}. These are the only modules in \kimeraMulti
 that involve communication between robots. 
 Fig.~\ref{fig:two_robot_dataflow} shows the data flow implemented by these modules. 
\subsection{Distributed Loop Closure Detection}
\label{sec:DLCD}

\begin{figure}[t]
	\centering
	\includegraphics[width=0.99\columnwidth,
	trim=0mm 0mm 0mm 0mm,clip]{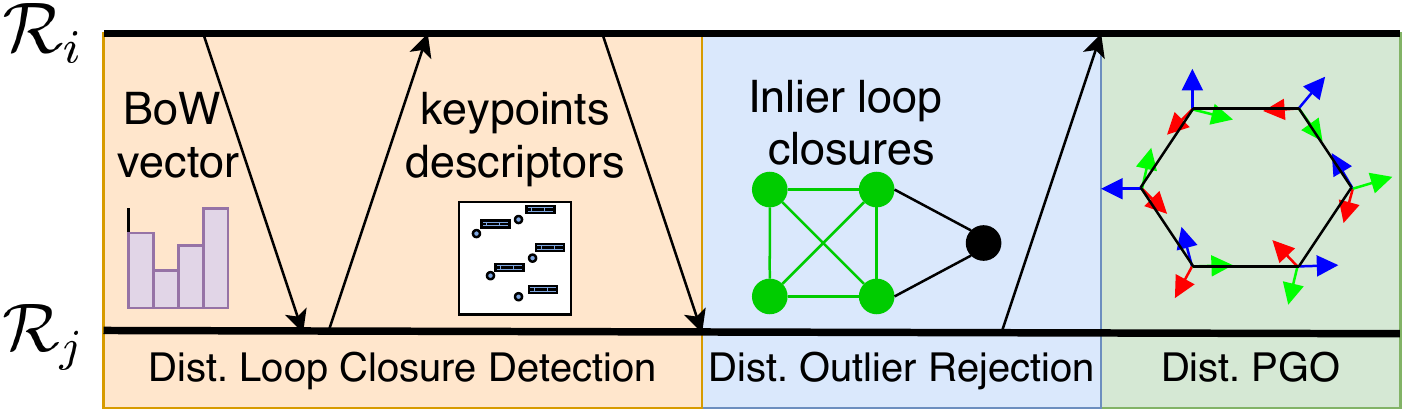}
	\vspace{-6mm}
	\caption{Communication protocol and data flow between pair of robots. \label{fig:two_robot_dataflow} }
	\vspace{-6mm}
\end{figure}

Each robot $\robot_i$ sends the BoW descriptors~\cite{Galvez12tro-dbow} of its keyframes to the other robot $\robot_j$.
When robot $\robot_j$ receives a new query from robot $\robot_i$, it searches among its own keyframes for candidate matches whose normalized visual similarity scores exceed a threshold ($\geq 0.5$ in our code).
When a potential loop closure is identified, the robots perform \emph{geometric verification} to estimate the relative transformation between the two matched keyframes. 
In our implementation, robot $\robot_j$ first requests the 3D keypoints and associated descriptors of the matched 
keyframe from robot $\robot_i$ (Fig.~\ref{fig:two_robot_dataflow}). 
Subsequently, robot $\robot_j$ computes putative correspondences by descriptor matching, and aligns the two sets of keypoints using Arun's method \cite{Arun87pami} with a 3-point RANSAC \cite{Fischler81}.  
If the final result contains sufficient inlier correspondences ($\geq 15$ inliers in our code), the loop closure is accepted and sent to the distributed outlier rejection module. 

\subsection{Distributed Outlier Rejection}
\label{sec:DPCM}

Similar to DOOR-SLAM~\cite{Lajoie20ral-doorSLAM}, robot $\robot_i$ then uses
PCM~\cite{Mangelson18icra} to reject spurious inter-robot loop closures. This is done by forming an
undirected graph  where nodes represent \emph{all} (new/old and inlier/outlier) inter-robot loop
closures between robots $\robot_i$ and $\robot_j$, and edges represent pairs of inter-robot
loop closures that are (pairwise)
``consistent''. The main idea behind PCM is that for any two inlier loop closures, 
composing the measurements along the cycle formed by the two
 loop closures and the (outlier-free) odometry in the pose graph must result in the identity
transformation. Based on this insight, PCM then classifies two inter-robot loop
closures as pairwise consistent if  
the composed noisy transformation along the cycle is sufficiently close to the identity.
We implement the consistency check and set the noise bounds as in~\cite[Sec.\ II-E]{Ebadi20icra-LAMP}.
A maximum clique in the PCM graph then corresponds to the largest set
of \emph{mutually consistent} inter-robot loop closures between the two robots, which is selected as the 
set of inliers for distributed PGO (Sec.~\ref{sec:DPGO}).

{\bf Incremental Approximate Maximum Clique.} The main challenge with PCM is that finding a maximum clique is
NP-hard. Therefore, here we propose an \emph{incremental} approximate solution, 
which \emph{updates} the maximum clique after each loop closure is detected, rather than 
\emph{initiating a
full search from scratch} as in~\cite{Rosinol20icra-Kimera,Lajoie20ral-doorSLAM,Pattabiraman15im-maxClique}.
Our main observation is that 
 after adding a new set of loop closures to the previous PCM
graph, exactly one of the following cases will occur: (i) the maximum clique
identified in the previous PCM graph remains a maximum clique for the new PCM
graph, which in turn implies that the inlier set remains unchanged; 
or (ii) there exists a larger clique in the new PCM graph, in which case the 
inlier set identified in the previous PCM graph must be replaced by a larger set. 
Therefore, our incremental search is based
on the insight that in the latter case, the new (larger) maximum clique
(inlier set) \emph{must} contain \emph{at least one of the new loop closures} (i.e.,
new vertices added to the PCM graph). This simple observation suggests the
following algorithm: rather than searching for the maximum
clique from scratch in the new PCM graph, it suffices to search for a largest
clique that {contains at least one of the new loop closures} and update the
inlier set \emph{only if} the size of this clique is larger than the size of the
maximum clique in the previous PCM graph. 

In practice, we perform this
\emph{restricted search} for each set of new loop closures and use the search heuristic
of~\cite{Pattabiraman15im-maxClique} to find the maximum clique in the subgraph induced by the new loop closures. We also leverage our knowledge of the size of the (approximate) maximum
clique in the previous graph to eliminate parts of the search space in our
restricted search that
\emph{cannot} lead to larger cliques; see the pruning strategies proposed in
\cite{Pattabiraman15im-maxClique}. Overall, our incremental PCM 
leverages the fact that the loop closures are added over time, to reduce the computational effort 
of the batch solutions in Kimera~\cite{Rosinol20icra-Kimera} and DOOR-SLAM~\cite{Lajoie20ral-doorSLAM}.
This is
crucial particularly because the size of the PCM graph continuously grows as new
loop closures are detected, making the computational cost of a batch solution 
grow unbounded over time.

\subsection{Distributed Pose Graph Optimization}
\label{sec:DPGO}
The odometry measurements and inlier loop closures returned by PCM are passed to {distributed pose graph optimization} to estimate the trajectories of all robots. 
We use the state-of-the-art Riemannian Block-Coordinate Descent (RBCD) solver \cite{tian2019distributed} as our PGO backend.
In short, RBCD solves the rank-restricted relaxation of PGO~\cite{Rosen18ijrr-sesync} (the default relaxation rank is set to 5) 
in a distributed fashion, and the solutions are subsequently projected to the special Euclidean group. 
Similar to the distributed Gauss-Seidel (DGS) method \cite{Choudhary17ijrr-distributedPGO3D}, RBCD only requires robots to exchange ``public poses'' (those that have inter-robot loop closures), and thereby preserves privacy (and saves communication effort) over the remaining poses. 
The main advantages of RBCD over DGS lies in the fact that it has provable convergence guarantees, while DGS uses an approximate decoupling. 
Moreover, BRCD can be used as an anytime algorithm, since each iteration is guaranteed to improve over the previous iterates by reducing the PGO cost function, while DGS requires completing rotation estimation before initiating pose estimation and is not a descent algorithm since
it is based on a Gauss-Newton scheme. %

\section{Local Mesh Optimization}
\label{sec:LMO}

\begin{figure}[t]
\centering
\begin{minipage}{0.5\textwidth}
\begin{tabular}{ccc}%
\hspace{1mm}
        \begin{minipage}{3.8cm}%
        \centering
        \includegraphics[width=\textwidth,trim=0mm 0mm 0mm 0mm,clip]{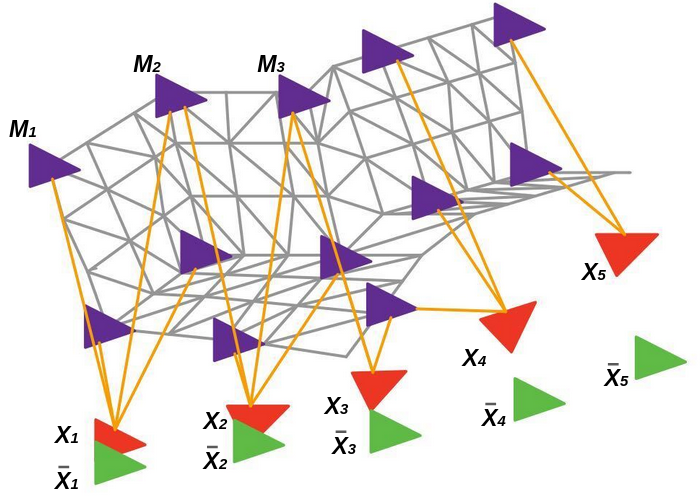} \\
        \scriptsize{(a) Undeformed mesh.} 
        \end{minipage}
    & \hspace{1mm}
        \begin{minipage}{3.8cm}%
        \centering
        \includegraphics[width=\textwidth,trim=0mm 0mm 0mm 0mm,clip]{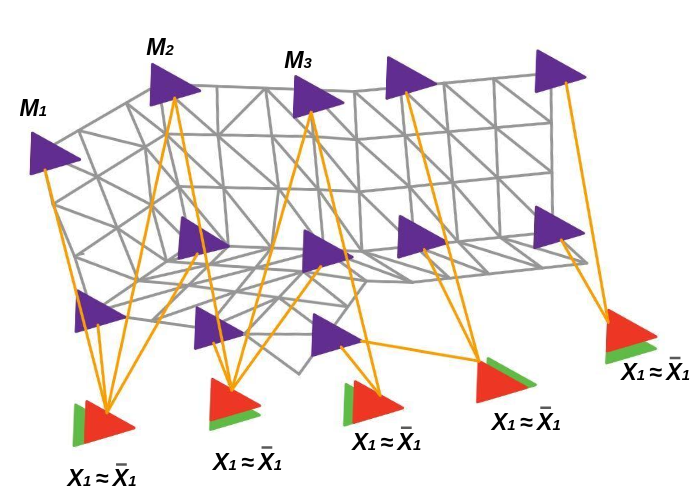} \\
        \scriptsize{(b) Deformed mesh. }
        \end{minipage}
    \end{tabular}
\end{minipage}
\caption{LMO deformation graph including mesh vertices (violet) and keyframe vertices (red). Edges connect two mesh vertices that 
are adjacent in the mesh (gray links), as well as mesh vertices with the keyframe vertices they are observed in (orange links). 
    \label{fig:deformation-graph}}
\vspace{-5mm}
\end{figure}

\kimeraSemantics builds the local 3D mesh at each robot $\robot_i$ using the pose estimates from \kimeraVIO.
 During this process, 
we keep track of the subset of 3D mesh vertices seen in each  keyframe from \kimeraVIO. 
This allow us to implement a 
\emph{local mesh optimization} (\LMO) approach to correct the 
mesh in response to changes in the keyframe poses --due to distributed PGO-- using \emph{deformation graphs}~\cite{Summer07siggraph-embeddedDeformation}.
\emph{Deformation graphs}~\cite{Summer07siggraph-embeddedDeformation} are a model from computer graphics 
that deforms a given mesh in order to anchor points in this mesh to user-defined locations while ensuring that the 
mesh remains locally rigid; deformation graphs are typically used for 3D animations, where one wants to animate a 3D object while 
ensuring it moves smoothly and without artifacts~\cite{Summer07siggraph-embeddedDeformation}.

In our \LMO approach, we first subsample the mesh from \kimeraSemantics 
to obtain a simplified mesh. Then, the vertices of this simplified mesh and the corresponding keyframe poses 
are added as \emph{vertices} in the deformation graph; we are going to refer to the corresponding vertices in the 
deformation graph 
as \emph{mesh vertices} and \emph{keyframe vertices}.
Moreover, we add two types of \emph{edges} to the deformation graph:
 \emph{mesh edges} (corresponding to pairs of mesh vertices sharing a face in the mesh),  
 and \emph{keyframe edges} (connecting a keyframe with the set of mesh vertices it observes).
For each mesh vertex $k$ in the deformation graph, we assign a transformation
$\MM_k = (\MR^M_k,\vt^M_k)$, where $\MR^M_k  \in \SOthree$  and  $\vt^M_k \in \Real{3}$; 
the pair $(\MR^M_k,\vt^M_k)$ defines a local coordinate frame, where
$\MR_k$ is initialized to the identity and $\vt_k$ is initialized to the position $\vg_k$ of the mesh vertex from \kimeraSemantics (\ie without accounting for loop closures). 
We also assign a pose $\MX_i = (\MR^x_k,\vt^x_k)$ to
 each keyframe vertex $i$ in the deformation graph. 
The pose is initialized to the pose estimates from \kimeraVIO.
Therefore, our goal is to adjust these poses (and the mesh vertex positions) to ``anchor'' 
the keyframe poses to the latest estimates from distributed PGO 
as shown in Fig~\ref{fig:deformation-graph}.

Denoting the optimized poses from distributed PGO as $\bar{\MX}_i$, 
and calling $n$ the number of keyframes in the trajectory and $m$ the total number of mesh vertices in the deformation graph, 
we compute updated poses $\MX_i$, $\MM_k$ of the vertices in the deformation graph 
by solving the following optimization:
\begin{align}
\label{eq:lmo}
		\argmin_{\substack{\MX_1,\ldots,\MX_n \in \SEthree \\ 
		\MM_1, \ldots,\MM_m \in \SEthree}} &
		\sum_{i=0}^n|| \MX_i \boxminus \bar{\MX}_i ||^2_{\Sigma_x} + \notag\\
	&\sum_{k=0}^m\sum_{l \in \mathcal{N}^M(k)}||\MR^M_k(\vg_l - \vg_k) + \vt^M_k - \vt^M_l||^2_{\Sigma} + \notag\\ 
	&\sum_{i=0}^n\sum_{l \in \mathcal{N}^M(i)}||\MR^x_i\widetilde{\vg}_{il} + \vt^x_i - \vt^M_l||^2_{\Sigma}
\end{align}
where, as before, $\vg_k$ denotes the non-deformed position of vertex $k$ in the deformation graph, 
$\widetilde{\vg}_{il}$ denotes the non-deformed position of vertex $l$ in the coordinate frame of keyframe $i$, 
 $\mathcal{N}^M(k)$ %
 denotes all the mesh vertices in the deformation graph connected to vertex $k$, 
 and $\boxminus$ denotes a tangent-space representation of the relative pose between  $\MX_i$ and $ \bar{\MX}_i$~\cite[7.1]{Barfoot17book}. 
Intuitively, the first term in the minimization~\eqref{eq:lmo} enforces (``anchors'') the poses of each keyframe $\MX_i$ 
to match the optimized poses $\bar{\MX}_i$ from distributed PGO.  
The second term enforces local rigidity of the mesh by minimizing the mismatch with respect to the non-deformed configuration
$\vg_k$.
The third term enforces local rigidity of  the relative positions between keyframes and mesh vertices by
 minimizing the mismatch with respect to the non-deformed configuration
in the local frame of pose $\MX_i$.
We optimize~\eqref{eq:lmo} using a Gauss-Newton method in GTSAM~\cite{gtsam}.

Since the deformation graph contains a subsampled version of the original mesh,
after the optimization, we retrieve the location of the remaining vertices as in~\cite{Summer07siggraph-embeddedDeformation}.
In particular, the positions of the vertices of the complete mesh are updated 
as affine transformations of nodes in the deformation graph: 
\begin{equation}
	\widetilde{\vv}_i = \sum_{j=1}^m w_j(\vv_i)[\MR^M_j(\vv_i - \vg_j) + \vt^M_j]
\end{equation}
where $\vv_i$ indicates the original vertex positions and $\widetilde{\vv}_i$ are the new deformed positions. 
The weights $w_j$ are defined as
\begin{equation}
		w_j(\vv_i) = \left(1 - ||\vv_i - \vg_j||/{d_{\text{max}}}\right)^2
\end{equation}
and then normalized to sum to one. Here $d_{\text{max}}$ is the distance to the $k + 1$ nearest node 
as described in \cite{Summer07siggraph-embeddedDeformation} (we set $k=4$).

Note that the \kimeraSemantics mesh also includes semantic labels, which remain untouched in the mesh deformation.

\section{Experiments}
\label{sec:experiments}

We evaluate  the trajectory and the metric-semantic mesh produced by \kimeraMulti 
in photo-realistic simulations and real data. Our results show that \kimeraMulti is an efficient, accurate, and robust
solution for distributed metric-semantic SLAM.

\subsection{Experimental setup}

{\bf Simulations.} We evaluate \kimeraMulti in two large-scale simulation scenarios, \camp (Fig.~\ref{fig:cover}) and \city (Fig.~\ref{fig:metric_semantic_reconstruction_city}), using the Unity-based simulator 
developed by the Army Research Laboratory
\emph{Distributed and Collaborative Intelligent Systems and Technology} (DCIST) Collaborative Research Alliance~\cite{dcist}.
Simulated robots are equipped with two RGB-D cameras and an IMU. 
The simulator provides ground-truth trajectories and map models (that we only use for benchmarking) as well as ground-truth 2D semantic segmentation that we use in \kimeraSemantics. 
In each scenario, we simulate 3 driving sequences (\ie 3 different robots) that we process with \kimeraMulti.

In addition, we use the Manhattan~\cite{Olson08thesis} synthetic PGO dataset to
evaluate the performance of our \emph{incremental} approximate maximum clique search
for outlier rejection.

{\bf Real data.}
We test \kimeraMulti on the \euroc dataset~\cite{Burri16ijrr-eurocDataset}.
\euroc includes data collected by a micro-aerial vehicle equipped with a grayscale stereo camera and IMU.
 We consider multiple \euroc sequences collected in the same environment (Vicon Room 1 and 2)
 and process them as they were sensor feeds from different robots, hence treating Vicon Room 1 and 2 (each with 3 sequences) as two multi-robot datasets.
\subsection{Incremental Approximate Maximum Clique}

\begin{figure}[t]
	\centering
	\subfloat[Run Time]{%
			\includegraphics[width=0.22\textwidth]{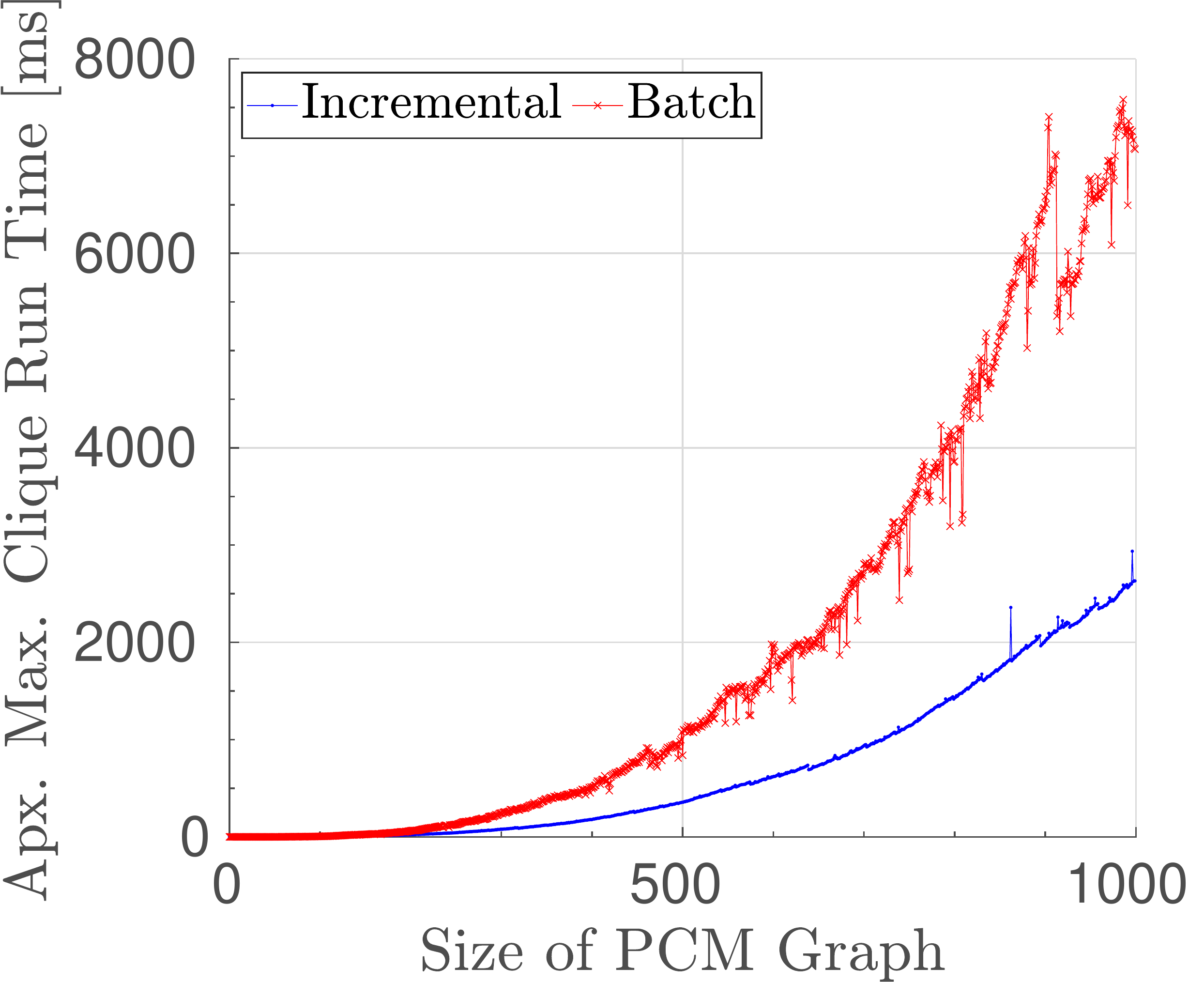}
	} ~
	\subfloat[Clique Size]{%
			\includegraphics[width=0.22\textwidth]{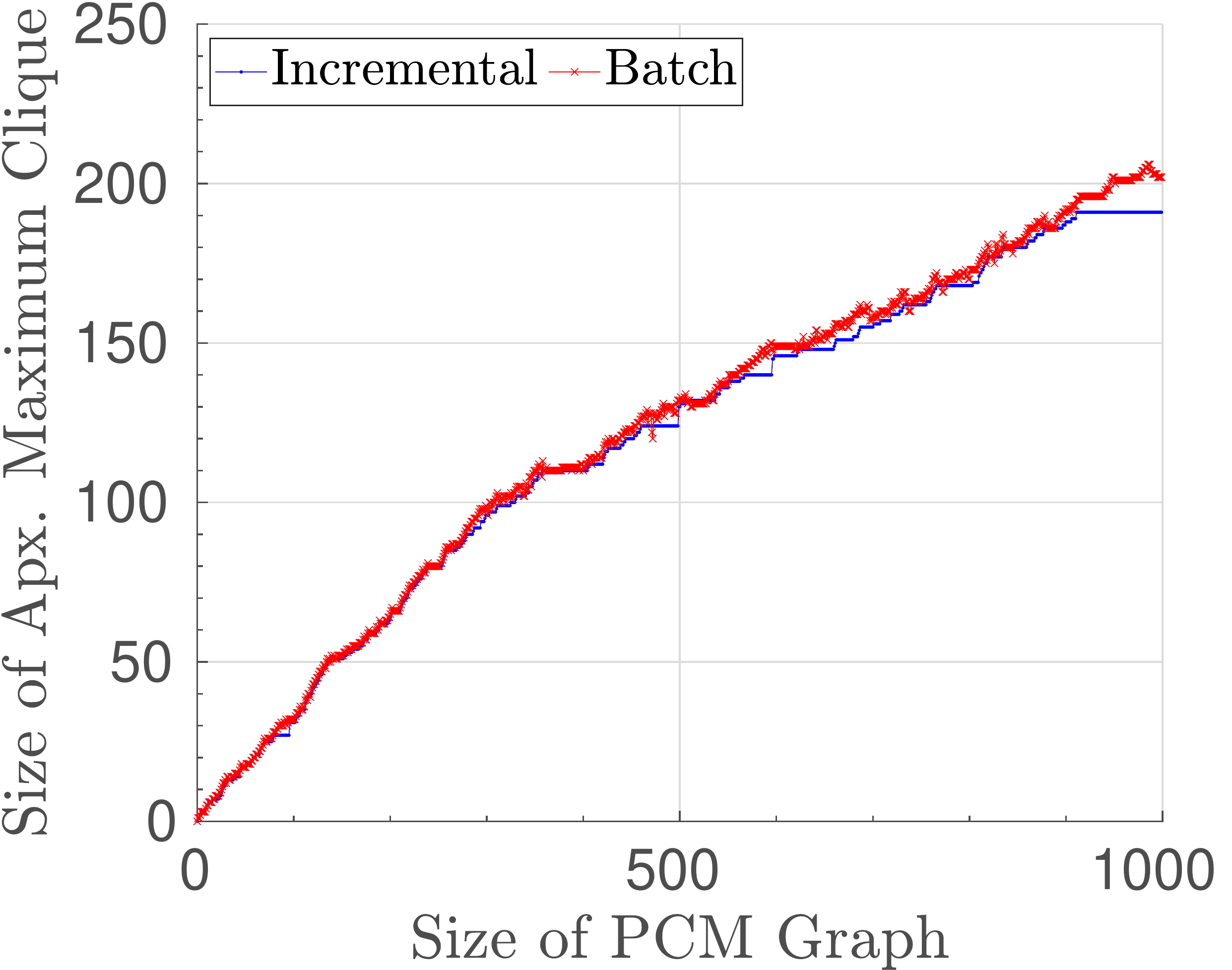}
	}
	\caption{
	(a) Approximate maximum clique search run
	times for our incremental PCM and the original (batch)
	PCM; (b) Size of the clique found by each method after adding
	each new loop closure.
	\label{fig:pcm}\vspace{-5mm}}
\end{figure}

We begin by comparing the run time of our new \emph{incremental} PCM (Sec.~\ref{sec:DPCM}) with the original (batch) PCM~\cite{Mangelson18icra} used 
in DOOR-SLAM~\cite{Lajoie20ral-doorSLAM} and Kimera~\cite{Rosinol20icra-Kimera}
for outlier rejection. To investigate how the run times scale
with the size of PCM graph in larger problems, we use the Manhattan dataset
\cite{Olson08thesis}. The original dataset has more than $2000$ loop closures. We added $1000$ randomly generated outlier loop closures
to this dataset using the tool provided by Vertigo~\cite{vertigoWebsite}. Loop
closures that were {\emph{not} consistent with the odometry were
immediately discarded~\cite[Sec.\ II.E]{Ebadi20icra-LAMP}.} The remaining loop
closures are added to the PCM
graph one by one. Upon adding each PCM vertex (loop closure), the edges of PCM graph
are updated based
on pairwise consistency checks, 
and then PCM (batch or incremental) is used to search for an approximate maximum clique using
the heuristic~\cite{Pattabiraman15im-maxClique}. %
Fig.~\ref{fig:pcm} shows (a) the run times and (b) the size of the approximate maximum clique
found by each approach for up to $1000$ PCM vertices. As expected, our
incremental approach significantly reduces the outlier rejection run time, while producing cliques (inlier sets) of comparable size.
Small differences between the clique sizes are caused by 
the underlying approximate search heuristic \cite{Pattabiraman15im-maxClique}.
The improvement in run time becomes more significant as the size of the PCM
graph (number of loop closures) increases. 

\subsection{Distributed Trajectory Estimation}
\begin{figure}[t]
	\centering
	\subfloat[City]{%
		\includegraphics[
		trim=25 170 55 200, clip,
		width=0.22\textwidth]
		{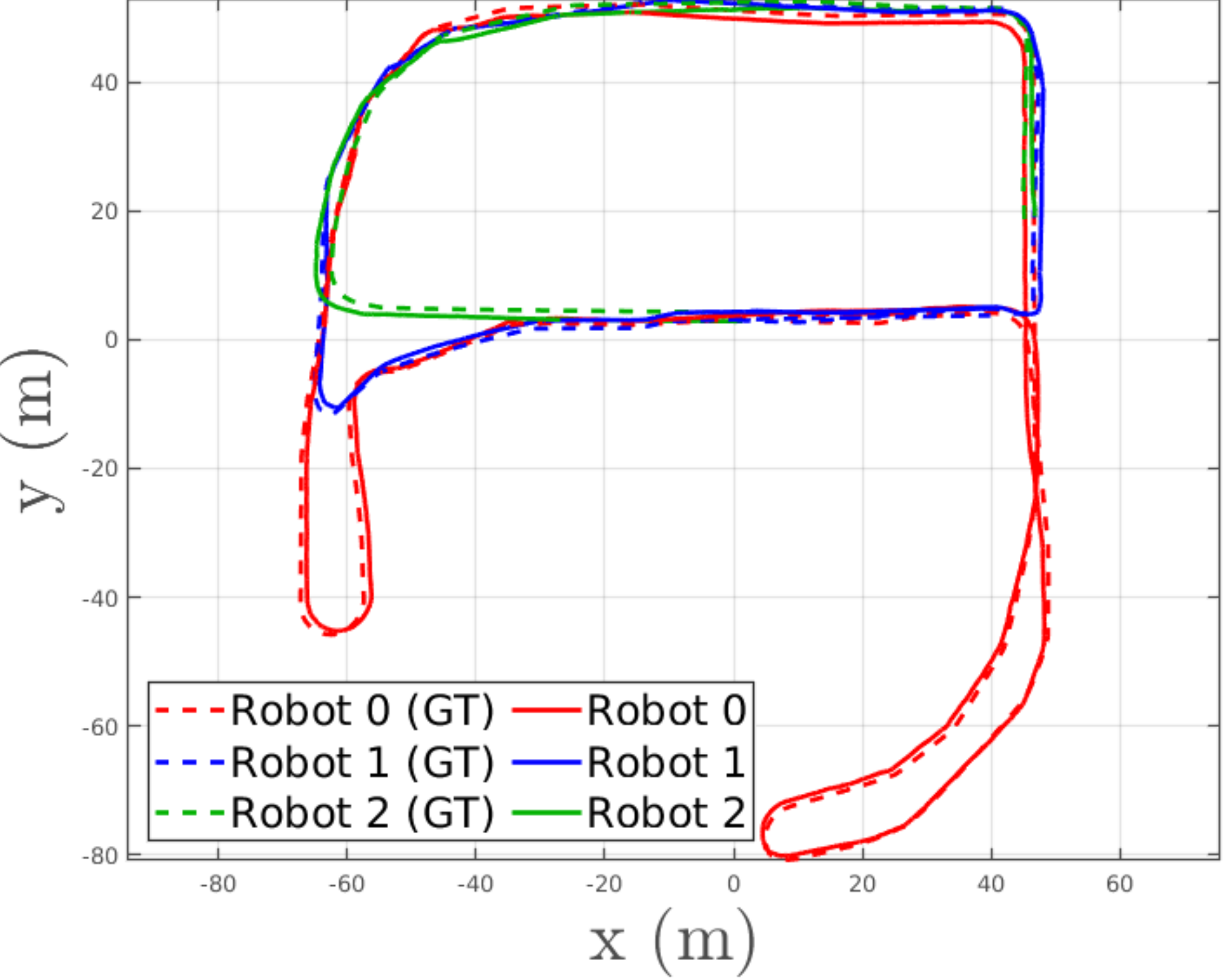}
	} ~
	\subfloat[Camp]{%
		\includegraphics[
		trim=25 170 55 200, clip,
		width=0.22\textwidth]
		{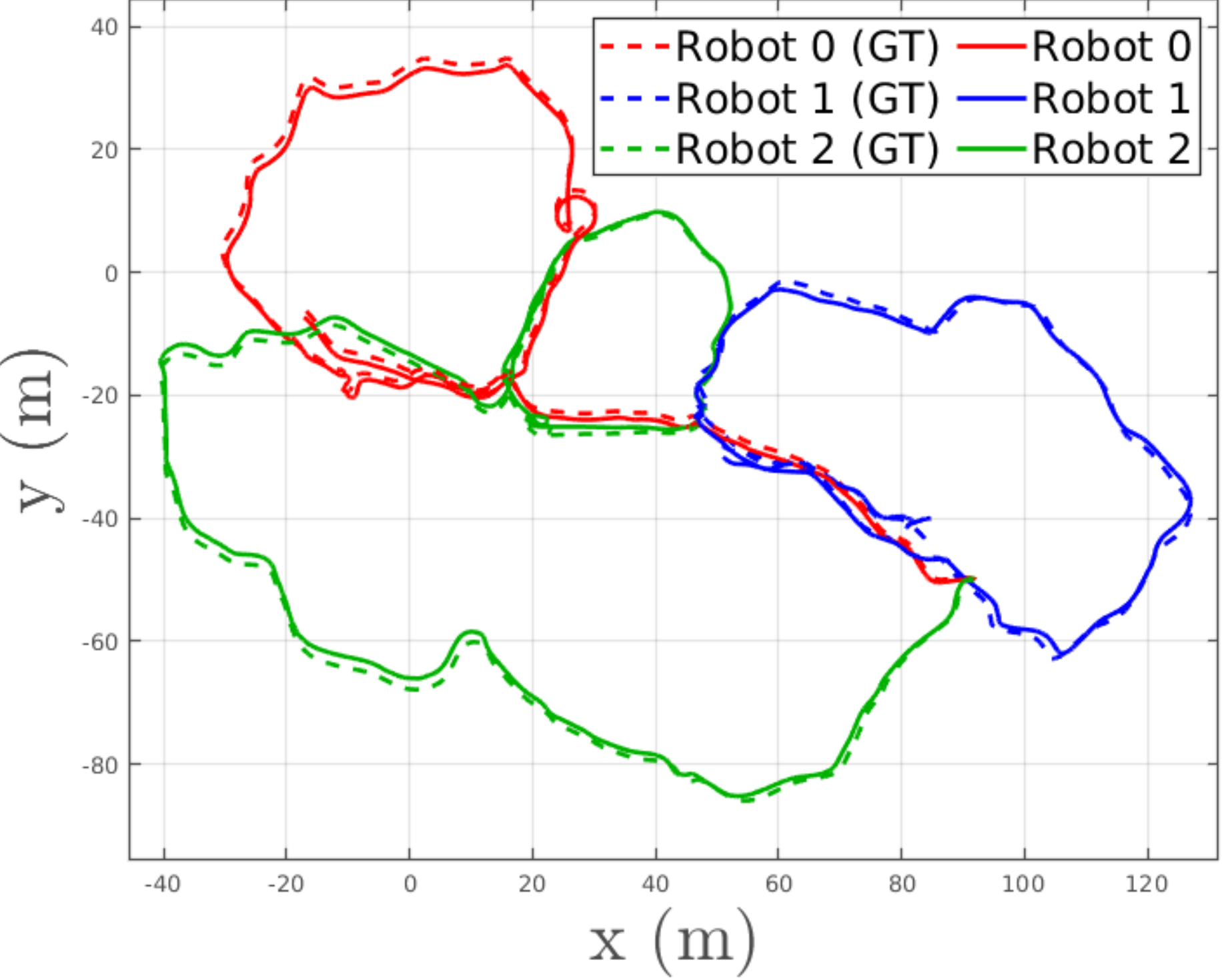}
	}
	\caption{Estimated trajectories in two simulation environments 	 (top-down view).
		Robots' trajectories are shown in different colors. 
		Corresponding ground truth are shown as dashed trajectories.\label{fig:trajectory_estimates} \vspace{-7mm}}
\end{figure}

In our experiments, we assume that robots are constantly in communication range.
Fig.~\ref{fig:trajectory_estimates} visualizes the trajectory estimates produced by \kimeraMulti in the two simulated scenes. 
Table~\ref{tab:traj_eval} shows the corresponding quantitative evaluations in simulation and on the \euroc datasets.
We report the absolute trajectory error (ATE), computed as the root mean squared error between the estimated and ground-truth trajectory. 
In addition, we also report the final PGO cost achieved by each of the compared techniques.
To demonstrate the flexibility of \RBCD as an anytime algorithm, we include the performance of an ``early stopped'' (label: ``\RBCD (ES)'') variant which terminates after 50 iterations. 
We compare the performance of \RBCD against a baseline method that transforms each robot's locally optimized trajectory to the global frame using a single inter-robot loop closure (label: ``local PGO''). 
In addition, we also compare against the two-stage distributed Gauss-Seidel method \cite{Choudhary17ijrr-distributedPGO3D} (label: ``DGS''). 
In our experiment, we run the first stage of DGS (rotation recovery) until convergence, and then run the second stage (pose recovery) for the same number of iterations as \RBCD. 
Lastly, we also include the centralized SE-Sync algorithm \cite{Rosen18ijrr-sesync} for reference. 
As shown in the table, \RBCD outperforms local PGO, which confirms the benefits of performing inter-robot 
loop closure detection and PGO. 
In simulation, \RBCD performs better compared to DGS in terms of both ATE and PGO cost, while the performances are similar on \euroc. %
On all datasets, the early stopped variant of \RBCD yields reasonably good trajectory estimates, which makes it a favorable choice in scenarios with run time constraints. 
Lastly, we observe that lower PGO costs generally translate to lower ATE, except on the Vicon Room 2 dataset in which the detected loop closures contain higher level of noise. 

\begin{table*}[t]
	\centering
	\caption{ \footnotesize
	Trajectory evaluation. We report the absolute trajectory error (ATE) in meters with respect to ground truth as well as the final PGO costs. 
	For each dataset, the best performing distributed method is highlighted in \textbf{bold} and SE-Sync performance is highlighted in \blue{blue}. }
	\label{tab:traj_eval}
	\renewcommand{\arraystretch}{1.3}
	\begin{tabular}{|c|c|c|c|c|c|c|c|c|c|c|}
		\hline
		&
		\multicolumn{5}{c|}{ATE {[}m{]}} &
		\multicolumn{5}{c|}{PGO cost} \\ \cline{2-11} 
		\multirow{-2}{*}{Dataset} &
		Local PGO &
		RBCD &
		RBCD (ES) &
		DGS &
		SE-Sync &
		Local PGO &
		RBCD &
		RBCD (ES) &
		DGS &
		SE-Sync \\ \hline
		City &
		6.09 &
		\textbf{2.38} &
		2.88 &
		2.62 &
		{\color[HTML]{0000FF} 1.46} &
		168532 &
		\textbf{1113.1} &
		3192.9 &
		1207.2 &
		{\color[HTML]{0000FF} 1047.3} \\ \hline
		Camp &
		2.81 &
		\textbf{2.28} &
		2.52 &
		2.58 &
		{\color[HTML]{0000FF} 2.28} &
		96695 &
		\textbf{140.45} &
		140.97 &
		171.03 &
		{\color[HTML]{0000FF} 140.32} \\ \hline
		Vicon Room 1 &
		0.455 &
		\textbf{0.317} &
		0.348 &
		0.346 &
		{\color[HTML]{0000FF} 0.308} &
		6451.9 &
		96.97 &
		138.18 &
		\textbf{91.97} &
		{\color[HTML]{0000FF} 90.61} \\ \hline
		Vicon Room 2 &
		0.473 &
		0.453 &
		\textbf{0.413} &
		0.503 &
		{\color[HTML]{0000FF} 0.453} &
		432.42 &
		\textbf{5.44} &
		7.19 &
		5.60 &
		{\color[HTML]{0000FF} 5.15} \\ \hline
	\end{tabular}
	\vspace{-2mm}
\end{table*}

Table~\ref{tab:resource_eval} shows the communication and computation costs of the distributed modules in our system.
We compare the amount of data transmission against a centralized SLAM system that transmits either raw images or extracted keypoints and descriptors to a base station. 
Overall, our system only uses 21\%-38\% of the total communication used by the centralized system that transmits image keypoints, which clearly demonstrates its communication efficiency. 
For computation, we report the total time and iterations spent by \RBCD 
and its early stopped variant.  For reference, we also include the run time of SE-Sync. 
As a first-order distributed method, \RBCD is slower than the centralized SE-Sync algorithm. 
Nevertheless, the early stopped variant of \RBCD is quite fast and still produces satisfactory solutions (see Table~\ref{tab:traj_eval}). 

\begin{table*}[t]
	\centering
	\caption{ \footnotesize
	Communication and computation usage of the proposed system. For communication, we show the total data transmitted in megabytes (MB) during place recognition (PR), geometric verification (GV), and distributed PGO (DPGO). We compare against centralized system that either transmits raw images or transmits detected keypoints and descriptors. For computation, we show the total number of iterations and run time for \RBCD and its early stopped (ES) variant. For reference, we also report the run time of the centralized SE-Sync method \cite{Rosen18ijrr-sesync}.    }
	\label{tab:resource_eval}
	\setlength{\tabcolsep}{4pt}
	\renewcommand{\arraystretch}{1.3}
	\begin{tabular}{|c|c|c|c|c|c|c|c|c|c|c|c|}
		\hline
		\multirow{2}{*}{Dataset} & \multirow{2}{*}{\# Poses} & \multirow{2}{*}{\# Edges} & \multicolumn{6}{c|}{Communication {[}MB{]}}    & \multicolumn{3}{c|}{Computation} \\ \cline{4-12} 
		&
		&
		&
		PR &
		GV &
		DPGO &
		Total &
		\begin{tabular}[c]{@{}c@{}}Centralized\\ (Image)\end{tabular} &
		\begin{tabular}[c]{@{}c@{}}Centralized\\ (Keypoints)\end{tabular} &
		\begin{tabular}[c]{@{}c@{}}RBCD\\ Iters / Time {[}sec{]}\end{tabular} &
		\begin{tabular}[c]{@{}c@{}}RBCD (ES)\\ Iters / Time {[}sec{]}\end{tabular} &
		\begin{tabular}[c]{@{}c@{}}SE-Sync \\ Time {[}sec{]}\end{tabular} \\ \hline
		City                     & 3238                      & 3428                      & 16.43 & 36.80 & 4.27 & 57.50 & 1989.4 & 149.48 & 500 / 34.90  & 50 / 2.76  & 2.23 \\ \hline
		Camp                     & 5189                      & 5411                      & 40.01 & 9.59  & 1.17 & 50.77 & 3188.1 & 242.66 & 128 / 33.16  & 50 / 6.49  & 4.34 \\ \hline
		Vicon Room 1             & 1690                      & 1730                      & 9.52  & 7.32  & 0.16 & 17.00 & 1223.9 & 73.76  & 95 / 3.02    & 50 / 1.53  & 0.33 \\ \hline
		Vicon Room 2             & 1526                      & 1544                      & 8.95  & 8.63  & 0.20 & 17.78 & 1105.1 & 69.35  & 173 / 5.84   & 50 / 1.72  & 0.47 \\ \hline
	\end{tabular}
	\vspace{-5mm}
\end{table*}

\subsection{Local Mesh Optimization}
\begin{figure}[t]
\vspace{-2mm}
	\centering
	\subfloat[Euroc]{%
		\includegraphics[
		trim=0 0 0 0, clip,
		width=0.23\textwidth]
		{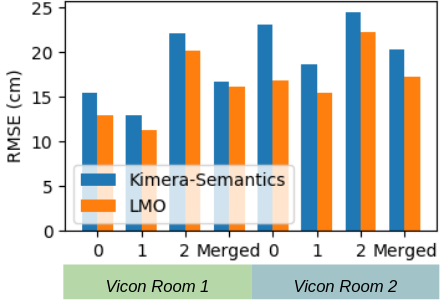}
	} ~
	\subfloat[Simulator]{%
		\includegraphics[
		trim=0 0 0 0, clip,
		width=0.23\textwidth]
		{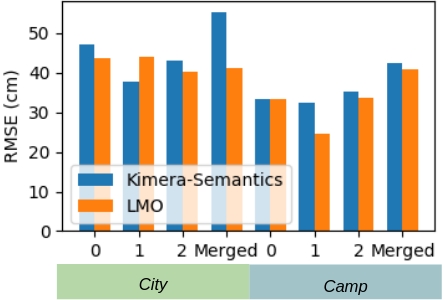}
	}
	\vspace{-1mm}
	\caption{Metric reconstruction evaluation. Mesh error (in centimeters) for the 3D meshes by \kimeraSemantics and \kimeraMulti's LMO.
	\label{fig:metric_accuracy} \vspace{-5mm}}
\end{figure}

\begin{figure}[t]
	\centering
	\includegraphics[width=0.99\columnwidth,
	trim=0mm 0mm 0mm 0mm,clip]{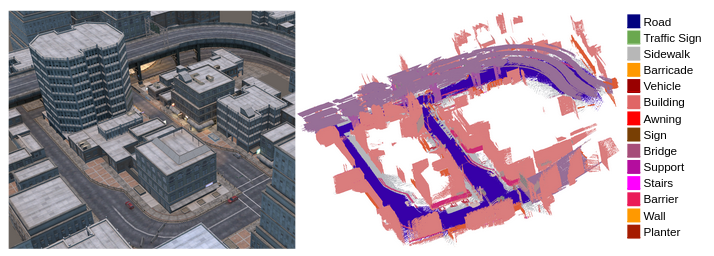}
	\vspace{-6mm}
	\caption{
Dense metric-semantic 3D mesh model generated by Kimera-Multi with three robots in the simulated \city scene. \label{fig:metric_semantic_reconstruction_city} }
	\vspace{-6mm}
\end{figure}

We use the ground-truth point clouds available in the \euroc Vicon Room 1 and 2 datasets, and the ground-truth mesh (and its semantic labels) available 
in the DCIST simulator to evaluate the accuracy of the 3D metric-semantic mesh by \kimeraSemantics and the impact of the local mesh optimization (LMO). %
For evaluation, the estimated and ground-truth meshes are sampled with a uniform density of $10^3~\text{points}/\text{m}^2$ as in~\cite{Rosinol20icra-Kimera}.
The resulting semantically-labeled point clouds are then registered using the ICP~\cite{Besl92pami} implementation
in \emph{Open3D}~\cite{Zhou18arxiv-open3D}. 
Then, we calculate the mean distance between each point in the ground-truth point cloud to its nearest neighbor in the estimated point cloud to obtain the metric accuracy of the 3D mesh. 
In addition, we evaluate the semantic reconstruction accuracy %
by calculating the percentage of correctly labeled points~\cite{Rosinol20icra-Kimera} relative to the ground truth 
using the correspondences given by ICP.
Fig.~\ref{fig:metric_accuracy} reports the metric accuracy  
of the individual meshes constructed by each robot as well as the merged 
global mesh, 
and Table~\ref{tab:semantic_eval_sim} shows the semantic reconstruction accuracy 
in the simulator (\euroc does not provide ground-truth semantics). In general, the metric-semantic mesh accuracy improves 
after LMO for both individual and merged 3D meshes, 
demonstrating the effectiveness of LMO in conjunction with our distributed trajectory optimization.
The dense metric-semantic meshes are shown in Fig.~\ref{fig:cover} and Fig.~\ref{fig:metric_semantic_reconstruction_city}.

\begin{table}[h]
	\caption{
	\footnotesize
	Semantic reconstruction evaluation. Semantic labels accuracy before and after correction by LMO in the DCIST simulator. \vspace{-2mm} }
	\label{tab:semantic_eval_sim}
	\centering
	\setlength{\tabcolsep}{1.6pt}
	\renewcommand{\arraystretch}{1.2}
	\begin{tabular}{|c|c|c|c|}
		\hline
		Dataset       & Robot ID & \kimeraSemantics (\%) &     LMO (\%)     \\ \hline
		              & 0       &    \bf{85.52}             &    \bf{85.52}    \\
		City          & 1       &    \bf{88.52}        &    85.65         \\ 
		              & 2       &    85.74             &    \bf{90.31}    \\ 
		              & Merged   &    67.91             &    \bf{83.93}    \\
		              \hline
		              & 0       &    \bf{95.83}        &    94.67         \\
		Camp          & 1       &    96.65             &    \bf{97.92}    \\ 
		              & 2       &    95.28             &    \bf{95.42}    \\
		              & Merged   &    94.22             &    \bf{95.13}    \\ 
		              \hline

	\end{tabular}
	\vspace{-3mm}
\end{table}

\section{Conclusion}
\label{sec:conclusion}

We present \kimeraMulti, the first fully distributed system that leverages a team of robots to build a dense
metric-semantic 3D mesh model of a large environment. 
\kimeraMulti combines recent advances in CPU-based metric-semantic mapping (\ie \kimera~\cite{Rosinol20icra-Kimera}), 
with state-of-the-art techniques for distributed pose graph optimization~\cite{tian2019distributed} (to which we add
an incremental maximum clique outlier rejection scheme),
and mesh deformation~\cite{Summer07siggraph-embeddedDeformation}.
We demonstrate \kimeraMulti in two photo-realistic large-scale simulations (\camp and \city environments) and on real data (\euroc).
\kimeraMulti is robust, efficient, and  builds accurate 3D metric-semantic meshes.
Future work includes 
(i) further enhancing \kimeraMulti with the Nesterov acceleration and solution verification techniques of~\cite{tian2019distributed},
(ii)~investigating distributed implementations of robust back-ends based on graduated non-convexity~\cite{Yang20ral-GNC}, 
and (iii)~unifying pose graph optimization and mesh deformation into a single optimization. 
\bibliographystyle{IEEEtran}

\begin{thebibliography}{10}
\providecommand{\url}[1]{#1}
\csname url@samestyle\endcsname
\providecommand{\newblock}{\relax}
\providecommand{\bibinfo}[2]{#2}
\providecommand{\BIBentrySTDinterwordspacing}{\spaceskip=0pt\relax}
\providecommand{\BIBentryALTinterwordstretchfactor}{4}
\providecommand{\BIBentryALTinterwordspacing}{\spaceskip=\fontdimen2\font plus
\BIBentryALTinterwordstretchfactor\fontdimen3\font minus
  \fontdimen4\font\relax}
\providecommand{\BIBforeignlanguage}[2]{{%
\expandafter\ifx\csname l@#1\endcsname\relax
\typeout{** WARNING: IEEEtran.bst: No hyphenation pattern has been}%
\typeout{** loaded for the language `#1'. Using the pattern for}%
\typeout{** the default language instead.}%
\else
\language=\csname l@#1\endcsname
\fi
#2}}
\providecommand{\BIBdecl}{\relax}
\BIBdecl

\bibitem{Rosinol20icra-Kimera}
A.~Rosinol, M.~Abate, Y.~Chang, and L.~Carlone, ``Kimera: an open-source
  library for real-time metric-semantic localization and mapping,'' in
  \emph{IEEE Intl. Conf. on Robotics and Automation (ICRA)}, 2020, arXiv
  preprint arXiv: 1910.02490,
  \linkToVideo{https://www.youtube.com/watch?v=-5XxXRABXJs&feature=youtu.be},
  \linkToCode{https://github.com/MIT-SPARK/Kimera},
  \linkToPdf{https://arxiv.org/pdf/1910.02490.pdf}.

\bibitem{Salas-Moreno13cvpr}
R.~F. Salas-Moreno, R.~A. Newcombe, H.~Strasdat, P.~H.~J. Kelly, and A.~J.
  Davison, ``{SLAM++}: Simultaneous localisation and mapping at the level of
  objects,'' in \emph{IEEE Conf. on Computer Vision and Pattern Recognition
  (CVPR)}, 2013.

\bibitem{McCormac17icra-semanticFusion}
J.~McCormac, A.~Handa, A.~J. Davison, and S.~Leutenegger, ``{SemanticFusion:
  Dense 3D Semantic Mapping with Convolutional Neural Networks},'' in
  \emph{IEEE Intl. Conf. on Robotics and Automation (ICRA)}, 2017.

\bibitem{Grinvald19ral-voxbloxpp}
M.~{Grinvald}, F.~{Furrer}, T.~{Novkovic}, J.~J. {Chung}, C.~{Cadena},
  R.~{Siegwart}, and J.~{Nieto}, ``{Volumetric Instance-Aware Semantic Mapping
  and 3D Object Discovery},'' \emph{{IEEE} Robotics and Automation Letters},
  vol.~4, no.~3, pp. 3037--3044, 2019.

\bibitem{Davison18arxiv-futureMapping}
A.~J. Davison, ``Futuremapping: The computational structure of spatial {AI}
  systems,'' 2018.

\bibitem{Rosinol20rss-dynamicSceneGraphs}
A.~Rosinol, A.~Gupta, M.~Abate, J.~Shi, and L.~Carlone, ``{3D} dynamic scene
  graphs: Actionable spatial perception with places, objects, and humans,'' in
  \emph{Robotics: Science and Systems (RSS)}, 2020,
  \linkToPdf{https://arxiv.org/pdf/2002.06289.pdf},
  \linkToMedia{http://news.mit.edu/2020/robots-spatial-perception-0715},
  \linkToVideo{https://www.youtube.com/watch?v=SWbofjhyPzI&feature=youtu.be}.

\bibitem{Cieslewski18icra}
T.~Cieslewski, S.~Choudhary, and D.~Scaramuzza, ``Data-efficient decentralized
  visual {SLAM},'' \emph{IEEE Intl. Conf. on Robotics and Automation (ICRA)},
  2018.

\bibitem{Choudhary17ijrr-distributedPGO3D}
S.~Choudhary, L.~Carlone, C.~Nieto, J.~Rogers, H.~Christensen, and F.~Dellaert,
  ``Distributed mapping with privacy and communication constraints: Lightweight
  algorithms and object-based models, accepted,'' \emph{Intl. J. of Robotics
  Research}, 2017, arxiv preprint: 1702.03435,
  \linkToPdf{http://arxiv.org/abs/1702.03435}
  \linkToWeb{https://www.economist.com/science-and-technology/2017/12/14/military-robots-are-getting-smaller-and-more-capable}
  \linkToCode{https://cognitiverobotics.github.io/distributed-mapper/}
  \linkToCode{https://github.com/uzh-rpg/dslam_open}
  \linkToCode{https://bitbucket.org/itzsid/object_slam/src/demo/cpp/ros/runDistributedORBSLAM_ROS_Robot.cpp}
  \linkToVideo{https://youtu.be/nXJamypPvVY}
  \linkToVideo{https://youtu.be/nYm2sSHuGjo}
  \linkToVideo{https://youtu.be/urZiIJK2IYk}
  \linkToVideo{https://youtu.be/-F6JpVmOrc0}.

\bibitem{Mangelson18icra}
J.~G. Mangelson, D.~Dominic, R.~M. Eustice, and R.~Vasudevan, ``Pairwise
  consistent measurement set maximization for robust multi-robot map merging,''
  in \emph{IEEE Intl. Conf. on Robotics and Automation (ICRA)}, 2018, pp.
  2916--2923.

\bibitem{Lajoie20ral-doorSLAM}
P.~Lajoie, B.~Ramtoula, Y.~Chang, L.~Carlone, and G.~Beltrame, ``{DOOR-SLAM:}
  distributed, online, and outlier resilient slam for robotic teams,''
  \emph{{IEEE} Robotics and Automation Letters ({RA-L})}, vol.~5, no.~2, pp.
  1656--1663, 2020, \linkToVideo{https://www.youtube.com/watch?v=h0bqURQlZGA},
  \linkToCode{https://github.com/MISTLab/DOOR-SLAM},
  \linkToPdf{https://arxiv.org/pdf/1909.12198.pdf}.

\bibitem{dcist}
{Army Research Laboratory}, ``{Distributed and Collaborative Intelligent
  Systems and Technology Collaborative Research Alliance (DCIST CRA)},''
  \url{https://www.dcist.org/}, 2020.

\bibitem{Oliva01ijcv}
A.~Oliva and A.~Torralba, ``Modeling the shape of the scene: a holistic
  representation of the spatial envelope,'' \emph{Intl. J. of Computer Vision},
  vol.~42, pp. 145--175, 2001.

\bibitem{Ulrich00icra}
I.~Ulrich and I.~Nourbakhsh, ``Appearance-based place recognition for
  topological localization,'' in \emph{IEEE Intl. Conf. on Robotics and
  Automation (ICRA)}, vol.~2, April 2000, pp. 1023 -- 1029.

\bibitem{Lowe99iccv}
D.~Lowe, ``Object recognition from local scale-invariant features,'' in
  \emph{Intl. Conf. on Computer Vision (ICCV)}, 1999, pp. 1150--1157.

\bibitem{Bay06eccv}
H.~Bay, T.~Tuytelaars, and L.~V. Gool, ``Surf: speeded up robust features,'' in
  \emph{European Conf. on Computer Vision (ECCV)}, 2006.

\bibitem{Sivic03iccv}
J.~Sivic and A.~Zisserman, ``Video google: a text re- trieval approach to
  object matching in videos,'' in \emph{Intl. Conf. on Computer Vision (ICCV)},
  2003.

\bibitem{Arandjelovic16cvpr-netvlad}
R.~{Arandjelovic}, P.~{Gronat}, A.~{Torii}, T.~{Pajdla}, and J.~{Sivic},
  ``{NetVLAD}: {CNN} architecture for weakly supervised place recognition,'' in
  \emph{IEEE Conf. on Computer Vision and Pattern Recognition (CVPR)}, 2016,
  pp. 5297--5307.

\bibitem{tian2019resource}
Y.~Tian, K.~Khosoussi, and J.~P. How, ``A resource-aware approach to
  collaborative loop closure detection with provable performance guarantees,''
  \emph{arXiv preprint arXiv:1907.04904}, 2019.

\bibitem{giamou2018talk}
M.~Giamou, K.~Khosoussi, and J.~P. How, ``Talk resource-efficiently to me:
  Optimal communication planning for distributed loop closure detection,'' in
  \emph{IEEE Intl. Conf. on Robotics and Automation (ICRA)}, 2018, pp. 1--9.

\bibitem{tian2018near}
Y.~Tian, K.~Khosoussi, M.~Giamou, J.~P. How, and J.~Kelly, ``Near-optimal
  budgeted data exchange for distributed loop closure detection,'' in
  \emph{Robotics: Science and Systems (RSS)}, 2018.

\bibitem{cieslewski2017efficient}
T.~Cieslewski and D.~Scaramuzza, ``Efficient decentralized visual place
  recognition using a distributed inverted index,'' \emph{{IEEE} Robotics and
  Automation Letters ({RA-L})}, vol.~2, no.~2, pp. 640--647, 2017.

\bibitem{van2018collaborative}
D.~Van~Opdenbosch and E.~Steinbach, ``Collaborative visual slam using
  compressed feature exchange,'' \emph{{IEEE} Robotics and Automation Letters
  ({RA-L})}, vol.~4, no.~1, pp. 57--64, 2018.

\bibitem{Tardioli2015}
D.~{Tardioli}, E.~{Montijano}, and A.~R. {Mosteo}, ``Visual data association in
  narrow-bandwidth networks,'' in \emph{IEEE/RSJ International Conference on
  Intelligent Robots and Systems (IROS)}, Sep. 2015, pp. 2572--2577.

\bibitem{Andersson08icra}
L.~Andersson and J.~Nygards, ``C-{SAM} : Multi-robot {SLAM} using square root
  information smoothing,'' in \emph{IEEE Intl. Conf. on Robotics and Automation
  (ICRA)}, 2008.

\bibitem{Kim10icra}
B.~Kim, M.~Kaess, L.~Fletcher, J.~Leonard, A.~Bachrach, N.~Roy, and S.~Teller,
  ``Multiple relative pose graphs for robust cooperative mapping,'' in
  \emph{IEEE Intl. Conf. on Robotics and Automation (ICRA)}, Anchorage, Alaska,
  May 2010, pp. 3185--3192.

\bibitem{Bailey11icra}
T.~Bailey, M.~Bryson, H.~Mu, J.~Vial, L.~McCalman, and H.~Durrant-Whyte,
  ``Decentralised cooperative localisation for heterogeneous teams of mobile
  robots,'' in \emph{IEEE Intl. Conf. on Robotics and Automation (ICRA)},
  Shanghai, China, May 2011, pp. 2859--2865.

\bibitem{Lazaro11icra}
M.~Lazaro, L.~Paz, P.~Pinies, J.~Castellanos, and G.~Grisetti, ``Multi-robot
  {SLAM} using condensed measurements,'' in \emph{IEEE Intl. Conf. on Robotics
  and Automation (ICRA)}, 2011, pp. 1069--1076.

\bibitem{Dong15icra}
J.~Dong, E.~Nelson, V.~Indelman, N.~Michael, and F.~Dellaert, ``Distributed
  real-time cooperative localization and mapping using an uncertainty-aware
  expectation maximization approach,'' in \emph{IEEE Intl. Conf. on Robotics
  and Automation (ICRA)}, Seattle, WA, May 2015, pp. 5807--5814.

\bibitem{Aragues11icra-distributedLocalization}
R.~Aragues, L.~Carlone, G.~Calafiore, and C.~Sagues, ``Multi-agent localization
  from noisy relative pose measurements,'' in \emph{IEEE Intl. Conf. on
  Robotics and Automation (ICRA)}, 2011, pp. 364--369.

\bibitem{tian2019distributed}
Y.~Tian, K.~Khosoussi, D.~Rosen, and J.~How, ``Distributed certifiably correct
  pose-graph optimization,'' \emph{arXiv preprint arXiv:1911.03721}, 2019.

\bibitem{tian2020asynchronous}
Y.~{Tian}, A.~{Koppel}, A.~S. {Bedi}, and J.~P. {How}, ``Asynchronous and
  parallel distributed pose graph optimization,'' \emph{IEEE Robotics and
  Automation Letters}, vol.~5, no.~4, pp. 5819--5826, 2020.

\bibitem{fan2020majorization}
T.~Fan and T.~Murphey, ``Majorization minimization methods to distributed pose
  graph optimization with convergence guarantees,'' \emph{arXiv preprint
  arXiv:2003.05353}, 2020.

\bibitem{cristofalo2020geod}
E.~Cristofalo, E.~Montijano, and M.~Schwager, ``Geod: Consensus-based geodesic
  distributed pose graph optimization,'' \emph{arXiv preprint
  arXiv:2010.00156}, 2020.

\bibitem{Cunningham10iros}
A.~Cunningham, M.~Paluri, and F.~Dellaert, ``{DDF-SAM}: Fully distributed slam
  using constrained factor graphs,'' in \emph{IEEE/RSJ Intl. Conf. on
  Intelligent Robots and Systems (IROS)}, 2010.

\bibitem{Cunningham13icra}
A.~Cunningham, V.~Indelman, and F.~Dellaert, ``{DDF-SAM} 2.0: Consistent
  distributed smoothing and mapping,'' in \emph{IEEE Intl. Conf. on Robotics
  and Automation (ICRA)}, Karlsruhe, Germany, May 2013.

\bibitem{Wang19arxiv}
\BIBentryALTinterwordspacing
W.~Wang, N.~Jadhav, P.~Vohs, N.~Hughes, M.~Mazumder, and S.~Gil, ``Active
  rendezvous for multi-robot pose graph optimization using sensing over
  {Wi-Fi},'' \emph{CoRR}, vol. abs/1907.05538, 2019. [Online]. Available:
  \url{http://arxiv.org/abs/1907.05538}
\BIBentrySTDinterwordspacing

\bibitem{SajadSaeedi2016MultipleRobotSL}
G.~S. Saeedi, M.~Trentini, M.~L. Seto, and H.~Li, ``Multiple-robot simultaneous
  localization and mapping: A review,'' \emph{J. of Field Robotics}, vol.~33,
  pp. 3--46, 2016.

\bibitem{Tchuiev20ral}
V.~{Tchuiev} and V.~{Indelman}, ``Distributed consistent multi-robot semantic
  localization and mapping,'' \emph{IEEE Robotics and Automation Letters},
  vol.~5, no.~3, pp. 4649--4656, 2020.

\bibitem{Tateno17cvpr-CNN-SLAM}
K.~{Tateno}, F.~{Tombari}, I.~Laina, and N.~{Navab}, ``{CNN-SLAM}: Real-time
  dense monocular slam with learned depth prediction,'' in \emph{IEEE Conf. on
  Computer Vision and Pattern Recognition (CVPR)}, 2017.

\bibitem{Lianos18eccv-VSO}
K.-N. Lianos, J.~L. Sch{\"o}nberger, M.~Pollefeys, and T.~Sattler, ``Vso:
  Visual semantic odometry,'' in \emph{European Conf. on Computer Vision
  (ECCV)}, 2018, pp. 246--263.

\bibitem{Dong17cvpr-XVIO}
J.~Dong, X.~Fei, and S.~Soatto, ``Visual-inertial-semantic scene representation
  for {3D} object detection,'' 2017.

\bibitem{Behley19iccv-semanticKitti}
J.~Behley, M.~Garbade, A.~Milioto, J.~Quenzel, S.~Behnke, C.~Stachniss, and
  J.~Gall, ``{SemanticKITTI: A Dataset for Semantic Scene Understanding of
  LiDAR Sequences},'' in \emph{Intl. Conf. on Computer Vision (ICCV)}, 2019.

\bibitem{Zheng19arxiv-metricSemantic}
L.~Zheng, C.~Zhu, J.~Zhang, H.~Zhao, H.~Huang, M.~Niessner, and K.~Xu, ``Active
  scene understanding via online semantic reconstruction,'' \emph{arXiv
  preprint:1906.07409}, 2019.

\bibitem{Tateno15iros-metricSemantic}
K.~{Tateno}, F.~{Tombari}, and N.~{Navab}, ``Real-time and scalable incremental
  segmentation on dense slam,'' in \emph{IEEE/RSJ Intl. Conf. on Intelligent
  Robots and Systems (IROS)}, 2015, pp. 4465--4472.

\bibitem{Li16iros-metricSemantic}
C.~{Li}, H.~{Xiao}, K.~{Tateno}, F.~{Tombari}, N.~{Navab}, and G.~D. {Hager},
  ``Incremental scene understanding on dense {SLAM},'' in \emph{IEEE/RSJ Intl.
  Conf. on Intelligent Robots and Systems (IROS)}, 2016, pp. 574--581.

\bibitem{McCormac183dv-fusion++}
J.~McCormac, R.~Clark, M.~Bloesch, A.~J. Davison, and S.~Leutenegger,
  ``Fusion++: Volumetric object-level {SLAM},'' in \emph{Intl. Conf. on 3D
  Vision (3DV)}, 2018, pp. 32--41.

\bibitem{Runz18ismar-maskfusion}
M.~Runz, M.~Buffier, and L.~Agapito, ``Maskfusion: Real-time recognition,
  tracking and reconstruction of multiple moving objects,'' in \emph{IEEE
  International Symposium on Mixed and Augmented Reality (ISMAR)}.\hskip 1em
  plus 0.5em minus 0.4em\relax IEEE, 2018, pp. 10--20.

\bibitem{Runz17icra-cofusion}
M.~R{\"u}nz and L.~Agapito, ``Co-fusion: Real-time segmentation, tracking and
  fusion of multiple objects,'' in \emph{IEEE Intl. Conf. on Robotics and
  Automation (ICRA)}.\hskip 1em plus 0.5em minus 0.4em\relax IEEE, 2017, pp.
  4471--4478.

\bibitem{Xu19icra-midFusion}
B.~Xu, W.~Li, D.~Tzoumanikas, M.~Bloesch, A.~Davison, and S.~Leutenegger,
  ``{MID-Fusion}: Octree-based object-level multi-instance dynamic slam,''
  2019, pp. 5231--5237.

\bibitem{Rosinol19icra-incremental}
A.~Rosinol, T.~Sattler, M.~Pollefeys, and L.~Carlone, ``Incremental
  visual-inertial 3d mesh generation with structural regularities,'' in
  \emph{IEEE Intl. Conf. on Robotics and Automation (ICRA)}, 2019.

\bibitem{Pattabiraman15im-maxClique}
B.~Pattabiraman, M.~M.~A. Patwary, A.~H. Gebremedhin, W.~K. Liao, and
  A.~Choudhary, ``Fast algorithms for the maximum clique problem on massive
  graphs with applications to overlapping community detection,'' \emph{Internet
  Mathematics}, vol.~11, no. 4-5, pp. 421--448, 2015.

\bibitem{Summer07siggraph-embeddedDeformation}
R.~W. Sumner, J.~Schmid, and M.~Pauly, ``Embedded deformation for shape
  manipulation,'' \emph{ACM SIGGRAPH 2007 papers on - SIGGRAPH '07}, 2007.

\bibitem{GarciaGarcia17arxiv}
A.~Garcia-Garcia, S.~Orts-Escolano, S.~Oprea, V.~Villena-Martinez, and
  J.~Garc{\'i}a-Rodr{\'i}guez, ``A review on deep learning techniques applied
  to semantic segmentation,'' \emph{ArXiv Preprint: 1704.06857}, 2017.

\bibitem{Galvez12tro-dbow}
D.~G\'alvez-L\'opez and J.~D. Tard\'os, ``Bags of binary words for fast place
  recognition in image sequences,'' \emph{IEEE Transactions on Robotics},
  vol.~28, no.~5, pp. 1188--1197, October 2012.

\bibitem{Quigley09icra-ros}
M.~Quigley, K.~Conley, B.~Gerkey, J.~Faust, T.~Foote, J.~Leibs, R.~Wheeler, and
  A.~Y. Ng, ``Ros: an open-source robot operating system,'' in \emph{ICRA
  workshop on open source software}, vol.~3, no. 3.2.\hskip 1em plus 0.5em
  minus 0.4em\relax Kobe, Japan, 2009, p.~5.

\bibitem{Arun87pami}
K.~Arun, T.~Huang, and S.~Blostein, ``Least-squares fitting of two 3-{D} point
  sets,'' \emph{{IEEE} Trans. Pattern Anal. Machine Intell.}, vol.~9, no.~5,
  pp. 698--700, sept. 1987.

\bibitem{Fischler81}
M.~Fischler and R.~Bolles, ``Random sample consensus: a paradigm for model
  fitting with application to image analysis and automated cartography,''
  \emph{Commun. ACM}, vol.~24, pp. 381--395, 1981.

\bibitem{Ebadi20icra-LAMP}
K.~Ebadi, Y.~Chang, M.~Palieri, A.~Stephens, A.~Hatteland, E.~Heiden,
  A.~Thakur, B.~Morrell, L.~Carlone, and A.~Aghamohammadi, ``{LAMP:}
  large-scale autonomous mapping and positioning for exploration of
  perceptually-degraded subterranean environments,'' in \emph{IEEE Intl. Conf.
  on Robotics and Automation (ICRA)}, 2020.

\bibitem{Rosen18ijrr-sesync}
D.~Rosen, L.~Carlone, A.~Bandeira, and J.~Leonard, ``{SE-Sync}: a certifiably
  correct algorithm for synchronization over the {Special Euclidean} group,''
  \emph{Intl. J. of Robotics Research}, 2018, accepted, arxiv preprint:
  1611.00128, \linkToPdf{https://arxiv.org/abs/1611.00128}.

\bibitem{Barfoot17book}
T.~Barfoot, \emph{State Estimation for Robotics}.\hskip 1em plus 0.5em minus
  0.4em\relax Cambridge University Press, 2017.

\bibitem{gtsam}
{F. Dellaert et al.}, ``{Georgia Tech Smoothing And Mapping (GTSAM)},''
  \url{https://gtsam.org/}, 2019.

\bibitem{Olson08thesis}
E.~Olson, ``Robust and efficient robotic mapping,'' Ph.D. dissertation,
  Massachusetts Institute of Technology, Cambridge, MA, USA, June 2008.

\bibitem{Burri16ijrr-eurocDataset}
M.~Burri, J.~Nikolic, P.~Gohl, T.~Schneider, J.~Rehder, S.~Omari, M.~Achtelik,
  and R.~Siegwart, ``The {EuRoC} micro aerial vehicle datasets,'' \emph{Intl.
  J. of Robotics Research}, 2016.

\bibitem{vertigoWebsite}
\BIBentryALTinterwordspacing
N.~S\"{u}nderhauf, ``Vertigo: Versatile extensions for robust inference using
  graph optimization.'' [Online]. Available:
  \url{http://openslam.org/vertigo.html}
\BIBentrySTDinterwordspacing

\bibitem{Besl92pami}
P.~J. Besl and N.~D. McKay, ``A method for registration of {3-D} shapes,''
  \emph{{IEEE} Trans. Pattern Anal. Machine Intell.}, vol.~14, no.~2, 1992.

\bibitem{Zhou18arxiv-open3D}
Q.-Y. Zhou, J.~Park, and V.~Koltun, ``{Open3D}: {A} modern library for {3D}
  data processing,'' \emph{arXiv:1801.09847}, 2018.

\bibitem{Yang20ral-GNC}
H.~Yang, P.~Antonante, V.~Tzoumas, and L.~Carlone, ``Graduated non-convexity
  for robust spatial perception: From non-minimal solvers to global outlier
  rejection,'' \emph{{IEEE} Robotics and Automation Letters ({RA-L})}, vol.~5,
  no.~2, pp. 1127--1134, 2020, arXiv preprint arXiv:1909.08605 (with
  supplemental material), \linkToPdf{https://arxiv.org/pdf/1909.08605.pdf},
  \award{ICRA Best paper award in Robot Vision}.

\end{thebibliography}

\end{document}